\documentclass[letterpaper]{article} 
\usepackage{aaai24}  
\usepackage{times}  
\usepackage{helvet}  
\usepackage{courier}  
\usepackage[hyphens]{url}  
\usepackage{graphicx} 
\usepackage{natbib}  
\usepackage{caption} 
\usepackage{algorithm}
\usepackage{algorithmic}
\usepackage{graphicx}
\usepackage{amsmath}
\usepackage{multirow}
\usepackage{amssymb}
\usepackage{arydshln}
\usepackage{xcolor}
\usepackage{dsfont}
\usepackage{newfloat}
\usepackage{listings}
\DeclareCaptionStyle{ruled}{labelfont=normalfont,labelsep=colon,strut=off} 
\floatstyle{ruled}
\newfloat{listing}{tb}{lst}{}
\floatname{listing}{Listing}
\title{Data Distribution Distilled Generative Model \\for Generalized Zero-Shot Recognition}
\author{
    Yijie Wang\textsuperscript{\rm 1},
    Mingjian Hong\textsuperscript{\rm 1},
    Luwen Huangfu\textsuperscript{\rm 2},
    Sheng Huang\textsuperscript{\rm 1}\thanks{Corresponding Author.}
}
\affiliations{
    \textsuperscript{\rm 1}Chongqing University\\
    \textsuperscript{\rm 2}Fowler College of Business, San Diego State University\\

    wangyj@stu.cqu.edu.cn,
    hmj@cqu.edu.cn,
    lhuangfu@sdsu.edu,
    huangsheng@cqu.edu.cn
}

\usepackage{bibentry}
\definecolor{gitcolor}{RGB}{255,22,255}

\begin{document}

\maketitle

\begin{abstract}
In the realm of Zero-Shot Learning (ZSL), we address biases in Generalized Zero-Shot Learning (GZSL) models, which favor seen data. To counter this, we introduce an end-to-end generative GZSL framework called D$^3$GZSL. This framework respects seen and synthesized unseen data as in-distribution and out-of-distribution data, respectively, for a more balanced model. D$^3$GZSL comprises two core modules: in-distribution dual space distillation (ID$^2$SD) and out-of-distribution batch distillation (O$^2$DBD). ID$^2$SD aligns teacher-student outcomes in embedding and label spaces, enhancing learning coherence. O$^2$DBD introduces low-dimensional out-of-distribution representations per batch sample, capturing shared structures between seen and unseen categories. Our approach demonstrates its effectiveness across established GZSL benchmarks, seamlessly integrating into mainstream generative frameworks. Extensive experiments consistently showcase that D$^3$GZSL elevates the performance of existing generative GZSL methods, underscoring its potential to refine zero-shot learning practices.The code is available at: https://github.com/PJBQ/D3GZSL.git

\end{abstract}

\section{Introduction}
Contemporary techniques for object classification \cite{he2016deep} are primarily rooted in supervised learning, mandating a substantial pool of labeled data.
Deep convolutional neural networks \cite{krizhevsky2017imagenet,tan2019efficientnet,xie2017aggregated} have elevated image classification performance within a predefined spectrum of categories, benefitting from an abundance of training samples.
However, real-world classification often involves a distribution skewed toward certain categories, resulting in limited or even absent samples for others.
The reliance of deep models on extensive data leads to suboptimal performance when labeled data is scarce \cite{wang2019survey}.
The emergence of zero-shot learning (ZSL) techniques \cite{lampert2009learning,palatucci2009zero} is prompted by real-world demands and technological progress.
ZSL endeavors to train a model capable of categorizing objects from unseen classes (target domain) by transferring knowledge gleaned from seen classes (source domain), employing semantic information as a channel between the two domains.
Traditional ZSL approaches involve test sets exclusively composed of samples from unseen classes, an unrealistic scenario that inadequately mirrors actual recognition conditions. In practical applications, seen class data samples typically outnumber those from unseen classes, making it imperative to concurrently identify samples from both categories, rather than restricting classification to unseen class samples alone. This more practical setting is known as generalized zero-shot learning (GZSL).

Generative methodologies  \cite{mishra2018generative,verma2018generalized,kong2022compactness} constitute a pivotal aspect of ZSL, enhancing its efficacy through data augmentation.
By generating samples for unseen classes, a GZSL problem can be transformed into a traditional supervised learning problem.
An illustrative example is f-CLSWGAN \cite{xian2018feature}, which utilizes the Wasserstein GAN \cite{arjovsky2017wasserstein} and classification loss to generate CNN features endowed with robust discriminatory attributes.
A complementary approach, f-VAEGAN-D2 \cite{xian2019f},  proposes a conditional generative model that combines the advantages of VAEs and GANs to generate more robust features.
An additional contribution is made by CE-GZSL \cite{han2021contrastive}, which harnesses both instance-level and class-level contrastive supervision to enhance the discriminative capabilities of the embedding space.

\begin{figure}[t]
  \centering
  \begin{minipage}[b]{0.45\textwidth}
    \includegraphics[width=\linewidth]{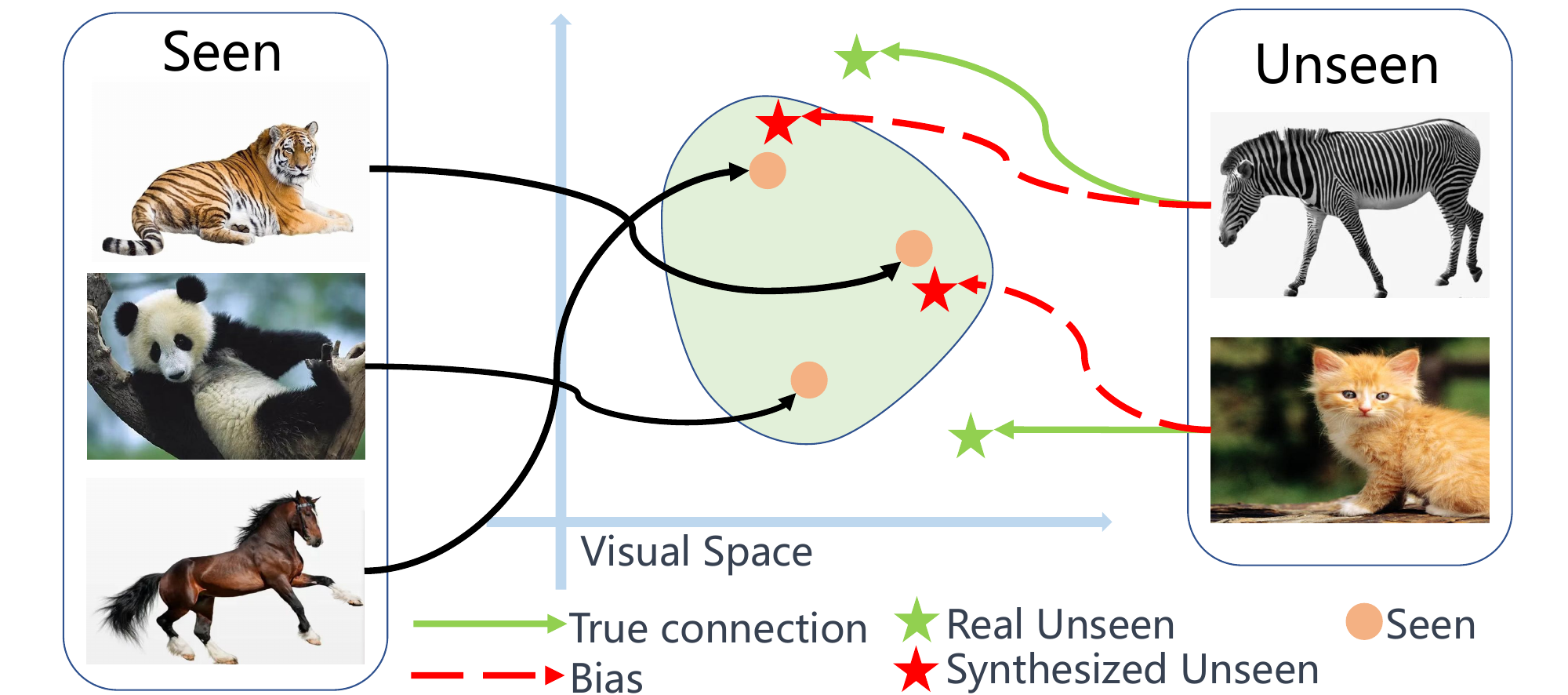}
    \caption{A schematic view of the bias concerning seen classes (source) in the visual space.}
    \label{fig:motivation1}
  \end{minipage}
  \hfill
  \begin{minipage}[b]{0.45\textwidth}
    \includegraphics[width=\linewidth]{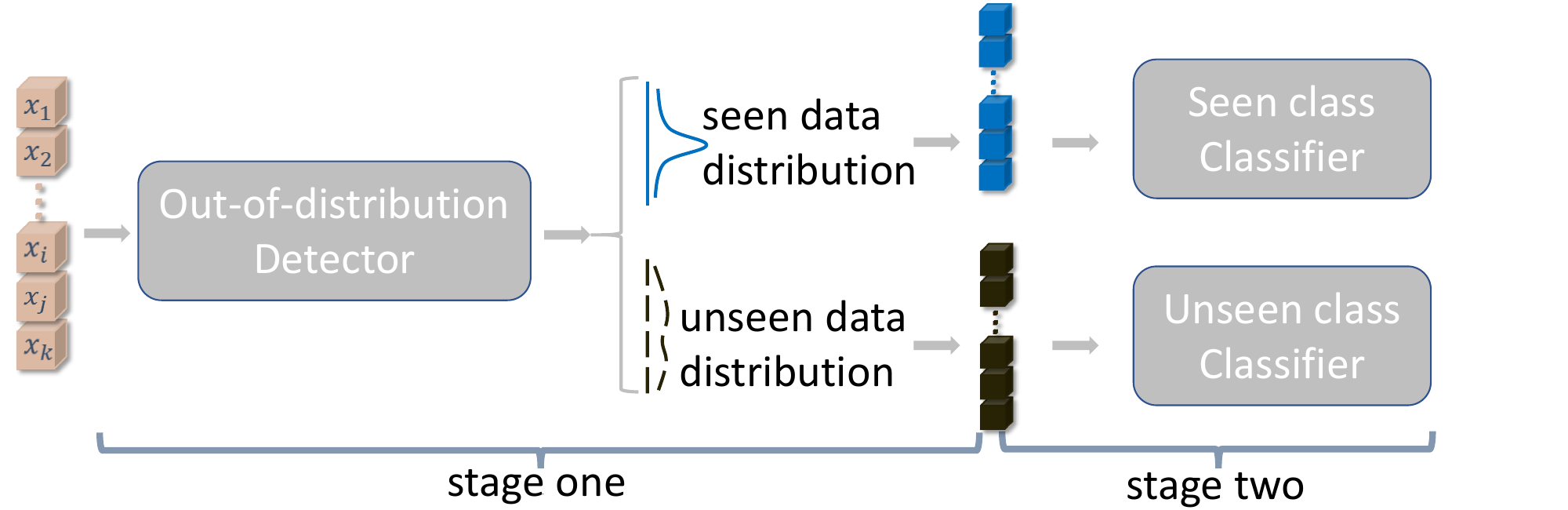}
    \caption{Two-stage classification method based on OOD detection. Stage one: OOD detector performs binary classification of the input data into seen and unseen categories. Stage two: Two expert classifiers separately classify the samples that the Out-Of-Distribution (OOD) detector identifies as seen and unseen categories.}
    \vspace{-20pt}
    \label{fig:ood_motivation}
  \end{minipage}
\end{figure}

Despite the impressive achievements of current generative methodologies \cite{xian2018feature,han2021contrastive}, a prevailing limitation surfaces. Most of these approaches, constrained by the scarcity of seen data, primarily concentrate on delineating the correlation between semantic knowledge and the available seen data. This dynamic naturally skews the generative model's inclination towards generating samples aligned with the seen data distribution, as depicted in Fig.\ref{fig:motivation1}.
Out-of-distribution (OOD) detection \cite{lee2018simple,sun2022out,sun2021react} aims to identify data samples that are abnormal or substantially distinct from the other available samples.
By reevaluating GZSL from the perspective of OOD detection, seen data can essentially be viewed as ID (In-Distribution) data, while the absent unseen data corresponds to OOD data. This perspective underscores that the bulk of mainstream generative techniques predominantly grapple with ID modeling, sidelining the crucial tasks of OOD modeling.
Fortunately, some researchers \cite{chen2020boundary,keshari2020generalized,mandal2019out} have recognized this limitation and have introduced OOD techniques to facilitate the incorporation of OOD insights.
For instance, \cite{mandal2019out} combines OOD detection techniques with generative methods to address the challenges encountered in ZSL.
As shown in Fig.\ref{fig:ood_motivation}, the training process of these methods is not end-to-end, and is usually divided into two steps.
A generative approach first trains to produce unseen samples, followed by training an OOD detector with both synthetic unseen and real seen samples. Simultaneously, two expert classifiers are trained separately on these generated unseen and authentic seen samples.
However, this non-end-to-end training strategy overlooks the potential benefits of OOD detection insights during the optimization of the generative model, ultimately neglecting to
model the data distribution across both seen and unseen classes.
In the inference phase, the OOD detector is applied to distinguish seen classes instances from those of the unseen classes and the domain expert classifiers (seen/unseen) are used to individually classify data samples.
Nonetheless, this two-stage classification approach can accrue errors, leading to suboptimal performance.

\begin{figure*}[t]
\centering
\includegraphics[width=\linewidth]{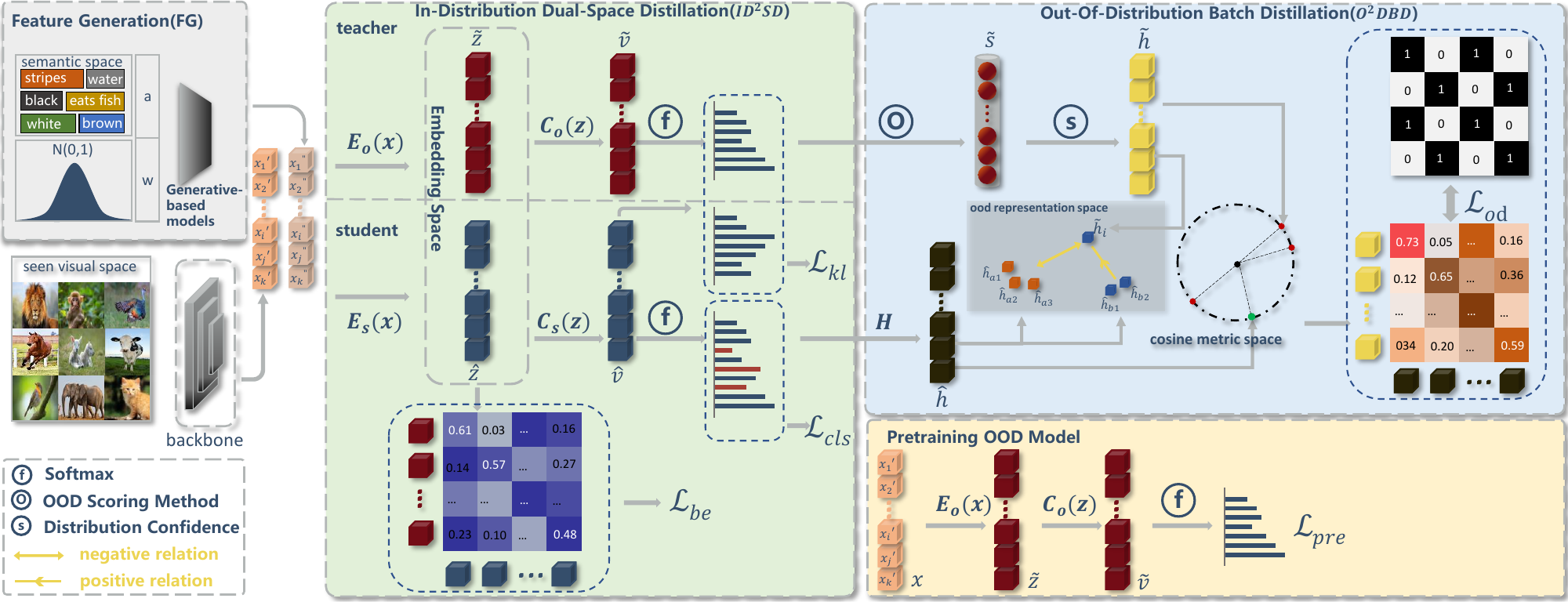}
\caption{The structure of our D$^3$GZSL framework. The FG is our baseline model, which is a generative ZSL method. In ID$^2$SD, we learn two embedding function $E_o$ and $E_s$ that map the visual samples $x$ into the embedding space as $z=E(x)$. $C_o$ and $C_s$ are the classifier networks of the teacher and student architectures, respectively. $f$ is a softmax function. In O$^2$DBD, $O$ is OOD scoring method. $H$ is a mapping function that maps the softmax probability of student network to the OOD representation embedding space. $S$ is the transformation of out-of-distribution detection scores into OOD representation space. }
\vspace{-15pt}
\label{fig:overview}
\end{figure*}

To address the above issues, we present a novel end-to-end knowledge distillation framework named Data Distribution Distilled Generative Zero-shot Learning (D$^3$GZSL) for mining and taking advantage of both ID and OOD knowledge of data, aiming to align the distribution of generated samples more closely with the distribution of actual samples.
By training a unified classifier for classification, we successfully circumvent the issue of error accumulation in two-stage classification encountered in \cite{chen2020boundary,mandal2019out}.
As shown in Fig.\ref{fig:overview}, our framework comprises a Feature Generation(FG), In-Distribution Dual-Space Distillation(ID$^2$SD), and Out-Of-Distribution Batch Distillation(O$^2$DBD).
The FG leverages the semantic relationships between seen and unseen classes to synthesize features for unseen classes.
ID$^2$SD is leveraged to align the outputs of teacher and student in both embedding and label spaces.
Logits-layer distillation enables our target network to model the probability distribution of in-distribution data (seen class) from the expert network. The feature-based distillation excavates the correlation between samples of different categories.
This dual-space approach could provide multifaceted guidance for the student model's learning process, thereby enhancing the precision of the distribution of seen category data.
In O$^2$DBD, we incorporate a low-dimensional representation to encode OOD information for each sample within a batch. Subsequently, we model the correlations among these low-dimensional OOD representations to capture the potentially shared inherent structures between seen and unseen categories.
This approach stimulates the model to grasp more nuanced, intricate features while simultaneously acquiring a diverse spectrum of features.

D$^3$GZSL can be applied to generally boost any generative GZSL models. The experiments on four benchmarks verify that D$^3$GZSL consistently improves the performances of three well known generative frameworks, such as a VAE-based model \cite{narayan2020latent} and two GAN-based models \cite{han2021contrastive,xian2018feature}.
Additionally, we have incorporated a novel generative GZSL method based on denoising diffusion model \cite{xiao2021tackling}.
The results also reveal that some generative GZSL approaches enhanced by our frameworks achieve promising performances compared with SOTA.

Our contributions encompass three key facets, each contributing uniquely to the advancement of GZSL:

(1) We introduce a novel generative framework for GZSL, named D$^3$GZSL, that operates in an end-to-end manner. Our framework incorporates the distilled knowledge emanating from both seen and unseen data distributions through the integration of OOD detection methodologies.

(2) Within our framework, ID$^2$SD plays a crucial role in harmonizing the outputs of both teacher and student networks across embedding and label spaces. Furthermore, in O$^2$DBD, we introduce a low-dimensional OOD representation for within each batch, subsequently modeling their interrelations under the guidance of class labels.

(3) Notably, our methodology is adaptable and can be seamlessly integrated into mainstream generative frameworks such as GAN, VAE, and the diffusion model.
This versatility allows our approach to enhance the capabilities of various existing generative GZSL methodologies.

\section{Related Work}
\textbf{Generalized Zero-Shot Learning.}
ZSL \cite{lampert2009learning,palatucci2009zero,xing2020robust,huang2020class} aims to train a model capable of classifying objects belonging to unseen classes (target domain) by leveraging knowledge acquired from seen classes (source domain), with the assistance of semantic information.
Early ZSL \cite{romera2015embarrassingly,kodirov2017semantic,liu2021goal,xu2022vgse} methods focused on learning an embedding space that connects the low-level visual features of seen classes to their corresponding semantic vectors.
Recently, generative methods have focused on learning a model to generate images or visual features for unseen classes, drawing on samples from seen classes and semantic representations of both classes.
\cite{xian2018feature} developed a conditional WGAN model incorporating a classification loss to generate visual features for unseen classes, a model known as f-CLSWGAN. Subsequent to this, the conditional Generative Adversarial Network (CGAN) has been integrated with various strategies to produce discriminative visual features for the unseen classes. TFGNSCS \cite{lin2019transfer} is an extended version of f-CLSWGAN, designed to take into account transfer information.
By putting more emphasis on
intra-class relationships but the inter-class structures, ICCE \cite{kong2022compactness}
can distinguish different classes with better generalization.
However, these generative-based models often create highly unconstrained visual features for unseen classes, which can result in synthetic samples that deviate significantly from the actual distribution of real visual features.
To address this issue, various strategies have been proposed, including calibrated stacking \cite{felix2019multi,chao2016empirical} and novelty detector \cite{atzmon2019adaptive,min2020domain}.
In this paper, we address this issue by incorporating out-of-distribution detection techniques into the task of generalized zero-shot recognition.

\textbf{Out-of-Distribution Detection.}
OOD detection \cite{hendrycks2016baseline,hendrycks2018deep,lee2017training,wang2022multi} is the task of detecting when a sample is drawn from a distribution different from the training data.
Some techniques \cite{huang2021importance,liang2017enhancing,liu2020energy,hendrycks2016baseline,lee2018simple,sun2021react} concentrate on developing score functions for OOD detection during the inference stage and can be easily implemented without modifying the model parameters.
The Maximum Softmax Probability (MSP) \cite{hendrycks2016baseline} approach uses the maximum prediction value from the model as the OOD score function.
Moreover, \cite{liu2020energy} proposes to replace the softmax function with the energy functions for OOD detection.
Recently, \cite{huang2021importance} has been introduced as an improvement in OOD detection, where it uses the similarity between the model's predicted probability distribution and a uniform distribution to attain state-of-the-art results.
Several pioneering works \cite{chen2020boundary,keshari2020generalized,mandal2019out} have utilized OOD to solve zero-shot learning tasks.
In \cite{chen2020boundary}, the boundary-based Out-of-Distribution (OOD) classifier learns a defined manifold for each seen class within a unit hyper-sphere, which serves as the latent space. Utilizing the boundaries of these manifolds along with their centers, unseen samples can be distinguished from the seen samples.

\textbf{Knowledge Distillation. }
Knowledge distillation \cite{hinton2015distilling} aims to train a more compact student network by replicating the behavior of a pretrained, complex teacher network. This knowledge can be response-based or feature-based.
Response-based \cite{chen2017learning,hinton2015distilling} knowledge typically refers to the neural response of the teacher model's final output layer, such as logits and bounding box offsets in object detection tasks.
Feature-based \cite{romero2014fitnets,zagoruyko2016paying,chen2021cross} knowledge from intermediate layers serves as a valuable extension of response-based knowledge, particularly for training thinner and deeper networks.
Specifically, \cite{zagoruyko2016paying} derived an "attention map" from the original feature maps for conveying knowledge.
Knowledge distillation is frequently utilized for model compression, fusion, or performance enhancement. In our approach, we amalgamate distillation learning with Out-of-Distribution detection techniques, aiming to align the distribution of generated samples more closely with the distribution of actual samples.

\section{Methodology}


\subsection{Problem Statement}
In GZSL, there are two separate sets of classes: $\mathcal{S}$ seen classes in $\mathcal{Y}_s$ and $\mathcal{U}$ unseen classes in $\mathcal{Y}_u$.
We define a training set $\mathcal{D}_{tr}=\{(x_i,y_i)\}^N_{i=1}$, $N$ is the number of seen images.
$x_i$ is a visual feature, $y_i$ is its class label in $\mathcal{Y}_s$, where we have $\mathcal{Y}_s\cap\mathcal{Y}_u=\varnothing$.
The test set $\mathcal{D}_{te}=\{x_i,y_i\}_{i=N+1}^{N+M}$ holds $M$ unlabeled instances.
The instances in $\mathcal{D}_{te}$ exclusively originate from both seen and unseen classes.
 Additionally, semantic embeddings (attributes) $\mathcal{A}=\{a_i\}_{i=1}^{S+U}$ for $S$ seen classes and $U$ unseen classes are also supplied.
 The attributes serve as the link between seen and unseen classes throughout the entire training process in GZSL settings.


\subsection{D$^3$-GZSL Framework}

As illustrated in Fig.\ref{fig:overview}, we present a data distribution distillation framework named D$^3$GZSL to generally boost generative models for achieving better GZSL performance.
D$^3$GZSL deems the real seen data and generated unseen data as the ID data and OOD data respectively.
It consists of FG, ID$^2$SD and O$^2$DBD.
FG module is used to generate the features for unseen classes. Here, it can be replaced any generative models.
ID$^2$SD aims to utilize the advantages of logits-layer and feature-based distillation to model the data distribution of seen classes in a detailed and accurate manner.
O$^2$DBD conducts an analysis from the OOD perspective with the objective of capturing the potentially shared inherent structures between both seen and unseen categories.
In this section, we will introduce these three modules in detail

\subsubsection{Feature Generation(FG). }
This module often consists of a generative GZSL approach.
Among these models, we have chosen four baseline models, including a VAE-based model \cite{narayan2020latent} and two GAN-based models \cite{han2021contrastive,xian2018feature}.
Furthermore, \cite{xiao2021tackling} introduced a novel denoising diffusion GAN, in which the denoising distributions are modeled with conditional GAN.
We incorporate their method\cite{xiao2021tackling} into zero-shot learning as one of our baseline models.
Building on this, We significantly modified the original image generation model for feature generation in GZSL tasks, detailed in the \textbf{supplementary materials}.
We use $\mathcal{L}_{gen}$ to represent the loss of these generative GZSL methods.
We utilize $G$ to symbolize generative models. $G$ is capable of transforming the semantic embedding $a$ and normal sampling $w$ into visual features $x$. We employ $x''$ to denote the generated features, thus expressing
\begin{equation}
    x''=G(a,w).
\end{equation}




\subsubsection{In-Distribution Dual-Space Distillation (ID$^2$SD). }
Our objective is to develop a generic classifier capable of distinguishing between both seen and unseen categories. We are able to reliably train a teacher network, utilizing only authentic samples.  We adopt the teacher-student network framework of dual-space distillation to build our ID$^2$SD model. The knowledge obtained from our reliable labels is distilled alongside the aspects corresponding to our unified classifier.
Similarly, at the feature level, we possess a trustworthy feature extractor $E_o$, with the aspiration that $E_s$ within our student network exhibits comparable capabilities. In fact, the distilled knowledge contains not only feature information but also mutual relations of data samples. Instead of measuring the similarity between the features directly through methods like MSE, we explore the correlation of samples within a batch matrix.
Our methodology consists of employing the embedding function $E_o$ along with the classifier $C_o$ to construct the teacher network. The student network architecture mirrors that of the teacher network, albeit with a distinction in that the classifier $C_s$ utilizes different dimensions. $C_s$ encompasses unified classifiers for both seen and unseen categories.
$x$ symbolizes the sample feature, constituting the input of our model, wherein $x^\prime$ stands for the real sample feature, and $x^{\prime\prime}$ denotes the generated sample feature. Where $\tilde{z} = E_o(x)$ and $\hat{z} = E_s(x)$, $z$ forms the features within our embedding space. $v$ represents the logical layer output of the network, expressed as $\tilde{v} = C_o(z)$ and $\hat{v} = C_s(z)$.
We employ $\phi$ to denote the composite function of $E_o$ and $C_o$, and use $\psi$ to symbolize the integration of $E_s$ and $C_s$, defining our teacher-student network as follows:
\begin{equation}
    \begin{split}
        \tilde{v}=\phi(x),\\
        \hat{v}=\psi(x).
    \end{split}
\label{equ:t_s_net}
\end{equation}

\textbf{Batch-Wise ID Embedding Identical Loss. }
We contemplate the uniformity of samples, along with the interrelation between sample features.
We initially construct the batch embedding similarity matrix $A$, wherein its element $a_{ij}$ as the cosine similarity between $\tilde{z}$ and $\hat{z}$, $a_{ij}=\frac{{\hat{z}_i}^T\tilde{z}_j}{{\lVert{\hat{z}_i}\rVert}_2{\lVert{\tilde{z}_j}\rVert}_2}$.
Our similarity matrix $A$ encodes the correlation between samples.
Following \cite{kong2022compactness}, we formulate a ground truth, designating $1$ for samples of identical classes and $0$ for those of disparate classes.
Thus, we metamorphose our task into a binary classification predicament. We constrain our loss calculations to utilize sample features exclusively from the seen category. Here, $N_a$ represents the number of samples belonging to the seen category within a batch. Our objective function is designed as follows:

\begin{equation}
\begin{split}
    \mathcal{L}_{be} = \frac{1}{N_a \times N_a} \sum_{i=1}^{N_a} \sum_{j=1}^{N_a} \bigl( \mathds{1}_{y_i=y_j} \log(\sigma(a_{ij})) \\
    + \mathds{1}_{y_i \ne y_j} \log(1-\sigma(a_{ij})) \bigr),
\end{split}
\label{equ:loss_be}
\end{equation}
where $\sigma(\cdot)$ is the sigmoid function.

\textbf{Instance-Wise ID Logit Identical Loss. }
The purpose of instance-wise ID logit identical loss is to align the outputs of the student and teacher networks in label space.
We initially extract the dimension corresponding to the seen category in $\hat{v}$, employing $\ddot{v}$ as a representation.
The $L_2$ normalized logits within the teacher and student network is articulated as follows: $v_{o}=\frac{\tilde{v}}{{\lVert \tilde{v}\rVert}_2}$ and $v_{s}=\frac{\ddot{v}}{{\lVert \ddot{v}\rVert}_2}$.
The teacher and student networks' probability distributions are obtained as follows:
\begin{equation}
\begin{split}
    {p_{o}}^{(k)}=\frac{\exp{({v_{o}}^{(k)})}}{\sum_{i=1}^S \exp{({{v_{o}}^{(i)})}}},k=1,2,...S,\\
    {p_{s}}^{(k)}=\frac{\exp{({v_{s}}^{(k)})}}{\sum_{i=1}^S \exp{({{v_{s}}^{(i)})}}},k=1,2,...S,
\end{split}
\end{equation}
where $S$ is the seen class number, $k$ is the class index. In pursuit of ensuring that projections from the identical seen class yield the same predicted probability, we introduce the normalized probability distillation loss, where $D_{KL}(p_{o}\lVert p_{s})$ denotes the $KL$ divergence between $p_{o}$ and $p_{s}$.
\begin{equation}
    \mathcal{L}_{kl}(p_{s},p_{o})=D_{KL}(p_{o}\lVert p_{s})=\sum_{k=1}^K {p_{o}}^{(k)}\log(\frac{{p_{o}}^{(k)}}{{p_{s}}^{(k)}}),
\end{equation}

\textbf{Classification Loss. }
Our framework constitutes an end-to-end trainable network, and the trained student network can be directly applied to GZSL inference. Beyond the dual-space distillation loss, we conduct supervised learning on the student network. The network's input encompasses the real features of the seen samples as well as the generated features of the unseen samples.
The cross-entropy loss is employed to supervise the classifier with the class labels:
\begin{equation}
    \mathcal{L}_{cls}(v,y)=-\sum_{i=1}^{S+U}y^{(i)}\log\frac{\exp(v^{(i)}/\tau_s)}{\sum_{k=1}^{S+U} \exp(v^{(k)}/\tau_s)},
\end{equation}
where $\tau_s>0$ is the temperature parameter, $S$ and $U$ respectively represent the number of seen classes and the number of unseen classes.
$y^{(i)}$ represents the i-th element in the true probability distribution, and $v^{(i)}$ constitutes the i-th element in the predicted vector.

\textbf{Total Loss of I$\boldsymbol{\mathrm{D}^2}$SD. }
The final optimization objective of our ID$^2$SD is formulated as:
\begin{equation}
    \mathcal{L}_{id}=\mathbb{E}[\mathcal{L}_{be}]+\mathbb{E}[\mathcal{L}_{kl}(p_{s},p_{o})]+\mathbb{E}[\mathcal{L}_{cls}(v,y)].
\end{equation}

\subsubsection{Out-of-Distribution Batch Distillation (O$^2$DBD).}


In O$^2$DBD, we introduce a low-dimensional representation to encode OOD information for each sample within a batch. And then we model the correlations among these low-dimensional OOD representations.
When unseen data serves as input, the teacher network generates a uniform distribution of equal probabilities across all seen categories, resulting in maximum entropy and inhibiting the model's ability to identify the input sample. Leveraging this distinctive feature of the OOD detection, we have constructed an out-of-distribution batch distillation network specifically for GZSL.

\textbf{OOD Logits. }
The score-based strategy is commonly used by many notable OOD detection methods.
The basic idea of this type of method is to use the model's output scores to determine whether an input comes from a known (i.e., seen during training) data distribution.
The model calculates a score for each input sample, indicating its confidence in the sample's category.
A key step in this method is determining a threshold $\gamma$ for the obtained scores, which is used to distinguish between ID and OOD data.
Diverging from traditional OOD detection methods, we have crafted an approach for confidence estimation, without the need to manually set a threshold value $\gamma$.
We obtain a low-level representation $\tilde{h}$, which encodes both the ID and OOD information.
$\tilde{h}$ consists of two elements. The first element is the ID confidence $c$, and the second element is the OOD confidence $\hat{c}$.
Through an OOD scoring method, we derive an OOD detection score $\tilde{s}$ (e.g., $\tilde{s}=\max($softmax$(F_o(x_i)))$ \cite{hendrycks2016baseline}).
We employ a learnable sigmoid function $\epsilon$ to compress the score $\tilde{s}$ into the range between 0 and 1, namely $c = \epsilon(\tilde{s})$.
We regard this as a probabilistic scenario, $\hat{c}=1-c$.
Then we learn a mapping function $H$ in the student network to map the softmax probability of student network to the OOD representation embedding space.
\begin{equation} \label{eq_s_confidence}
    \hat{h}=H(\psi(x)).
\end{equation}

\textbf{Batch-Wise OOD Logits Identical Loss. }
We employ the identical loss function as expressed in Eq.\ref{equ:loss_be}, utilizing the confidence label $\tilde{h}$, to construct supervised learning in a low-dimensional space that encodes both ID and OOD information.
We first create an OOD representation similarity matrix $B$, whose element $b_{ij}$ is derived by calculating the cosine similarity between $\tilde{h}$ and $\hat{h}$,  $b_{ij}=\frac{{\hat{h}_i}^T\tilde{h}_j}{{\lVert{\hat{h}_i}\rVert}_2{\lVert{\tilde{h}_j}\rVert}_2}$.
Our aim is to augment the confidence of the same class within the student and teacher networks while minimizing the confidence of disparate classes.
The deviation from Eq.\ref{equ:loss_be} lies in that our input includes both seen and unseen samples, and $N_b$ symbolizes the OOD representation numbers within a batch.
Our objective function is designed as follows:
\begin{equation}
\begin{split}
    \mathcal{L}_{od}=\frac{1}{{N_b}\times {N_b}}\sum_{i=1}^{N_b} \sum_{j=1}^{N_b} {(\mathds{1}_{y_i=y_j}\log(\sigma(b_{ij}))}\\
    +\mathds{1}_{y_i\ne y_j}\log(1-\sigma(b_{ij}))).
\end{split}
\end{equation}

\begin{table*}[ht]
\centering
\small
\begin{tabular}{c|ccc|ccc|ccc|ccc}
\hline
\multirow{2}{*}{Method} & \multicolumn{3}{c|}{\textbf{AWA1}} & \multicolumn{3}{c|}{\textbf{AWA2}} & \multicolumn{3}{c|}{\textbf{CUB}} & \multicolumn{3}{c}{\textbf{FLO}}\\
& U & S & H & U & S & H &U & S & H &U & S & H\\
\hline
cycle-CLSWGAN \cite{felix2018multi} & 56.9& 64.0 &60.2& -& -& - &45.7& 61.0& 52.3 &59.2 &72.5& 65.1\\
CADA-VAE \cite{schonfeld2019generalized} & 57.3& 72.8 &64.1 &55.8 &75.0 &63.9& 51.6 &53.5 &52.4& -& - &- \\
LisGAN \cite{li2019leveraging} & 52.6 & 76.3 & 62.3 & - & - & - & 46.5 & 57.9 & 51.6 & 57.7 & 83.8 &68.3 \\
IZF \cite{shen2020invertible} & 61.3& \textbf{80.5} & \underline{69.6} &60.6&77.5& 68.0 &52.7& 68.0 &59.4&-&-&-\\
 SE-GZSL \cite{kim2022semantic} & 61.3 & \underline{76.7} & 68.1 & 59.9 & \underline{80.7} & 68.8 & 53.1 & 60.3 & 56.4 & -& - & - \\


TDCSS \cite{feng2022non} & 54.4 & 69.8 & 60.9 & 59.2 & 74.9 & 66.1 & 44.2 & 62.8 & 51.9 & 54.1 & 85.1 & 66.2\\


DUET \cite{chen2023duet} & - & - & - & \underline{63.7} & \textbf{84.7} & \textbf{72.7} & 62.9 & \textbf{72.8} & \underline{67.5}  & - & - & - \\

GKU \cite{guo2023graph}  & - & - & - & - & - & - & 52.3 & \underline{71.1} & 60.3 & - & - & - \\
\hline
f-CLSWGAN \cite{xian2018feature} & 57.9 & 61.4 & 59.6 & - & - & - & 43.7 & 57.7 & 49.7 & 59.0 & 73.8 & 65.6\\
\textbf{f-CLSWGAN+D$^3$GZSL} & 57.1 & 69.8 & 62.8 & - & - & - & 52.3 & 61.5 & 56.5 & 61.1 & \textbf{86.7} & 71.7\\
\hline
TF-VAEGAN \cite{narayan2020latent} & - & - & - & 59.8 & 75.1 & 66.6 & 52.8 & 64.7 & 58.1 & 62.5 & 84.1 & 71.7\\
\textbf{TF-VAEGAN+D$^3$GZSL} & - & - & - & 60.2 & 74.9 & 66.8 & 57.3 & 64.5 & 60.7 & 65.6 & 81.4 & 72.7 \\
\hline

DDGAN & 58.1 & 63.5 & 60.7 & 61.7 & 68.4 & 64.9 & 47.5 & 61.5 & 53.6 & 61.1 & \underline{85.2} & 71.2\\
\textbf{DDGAN+D$^3$GZSL} & 59.5 & 68.3 & 63.6 & 62.9 & 67.7 & 65.2 & 54.2 & 59.7 & 56.8 & 63.4 & 82.1 & 71.5\\
\hline
 CE-GZSL \cite{han2021contrastive} & \underline{65.3} & 73.4 & 69.1 & 63.1 & 78.6 & 70.0 & \underline{63.9} & 66.8 & 65.3 & \textbf{69.0} & 78.7 & \underline{73.5}\\
 \textbf{CE-GZSL+D$^3$GZSL} & \textbf{65.7} & 76.2 & \textbf{70.5} & \textbf{64.6} & 76.7 & \underline{70.1} & \textbf{66.7} & 69.1 & \textbf{67.8} & \underline{68.6} & 80.9 & \textbf{74.2}\\
\hline
\end{tabular}
\caption{Comparisons with state-of-the-art GZSL methods and baseline generative ZSL methods. D$^3$GZSL represents the use of our D$^3$GZSL framework based on the baseline model. $U$ and $S$ are the Top-1 accuracy of the unseen and seen classes, respectively. $H$ is the harmonic mean of $U$ and $S$.  The best and second best results are marked in bold and underline, respectively.}
\vspace{-10pt}
\label{table:sota}
\end{table*}

\subsection{Model Optimization}
\textbf{Training. }
Our framework first undergoes a preprocessing phase in which the teacher model is trained on real seen samples. The teacher model then extracts knowledge, either in the form of logits or as intermediate features. This extracted knowledge is subsequently used to guide the training of the student model during the distillation process.
The cross-entropy loss in the preprocessing phase is as follows:
\begin{equation}
    \mathcal{L}_{pre}(x,y)=-\sum_{i=1}^{S}y^{(i)}\log\frac{\exp({\phi(x)}^{(i)})/\tau_o)}{\sum_{k=1}^S \exp({\phi(x)}^{(k)})/\tau_o)}.
\end{equation}

After completing the pre-training of the OOD detection model, we jointly train the FG, ID$^2$SD and O$^2$DBD end-to-end.
We utilize the real seen samples $x'$ and the unseen samples $x''$ generated by the FG as inputs for the ID$^2$SD module. Then, we calculate the OOD confidence labels from the output of the teacher network, and map the softmax probability of student network to the OOD representation embedding space.
Thus, the total loss of D$^3$GZSL is formulated as:
\begin{equation}
    \min_{G,E_s,C_s,H} \mathcal{L}_{gen} + \lambda(\mathcal{L}_{id} + \mathcal{L}_{od}),
\end{equation}
where $\lambda$ is the hyper-parameters indicating the effect
of $\mathcal{L}_{id}$ and $\mathcal{L}_{od}$ towards the generator.

\textbf{Inference. }
We no longer train a separate classifier for classification.
Once the training is completed, we map the $\mathcal{D}_{te}=\{x_i,y_i\}_{i=N+1}^{N+M}$ to the embedding space using the embedding function $E_s$ of the student network. Then, we employ the classifier $C_s$ to predict the class label $\hat{y}$:
\begin{equation}
    \begin{split}
        \hat{y} = \arg\max_i \frac{{\exp({\psi(x)}}^{(i)})}{\sum_{k=1}^{S+U}\exp({{\psi(x)}^{(k)}})}.
    \end{split}
\end{equation}

\begin{table*}[ht]
\centering
\small
\begin{tabular}{c|ccc|ccc|ccc|ccc|ccc}
\hline
\multirow{2}{*}{Method} & \multirow{2}{*}{IV-TS} & \multirow{2}{*}{TS} & \multirow{2}{*}{Ours} & \multicolumn{3}{c|}{\textbf{AWA1}} & \multicolumn{3}{c|}{\textbf{AWA2}} & \multicolumn{3}{c|}{\textbf{CUB}} & \multicolumn{3}{c}{\textbf{FLO}}\\
&&&&U & S & H & U & S & H &U & S & H &U & S & H\\
\hline
\multirow{3}{*}{f-CLSWGAN}
& \checkmark & & & 67.4 & 88.9 & - & - & - & - & 56.1 & 70.0 & - & 65.2 & 87.4 & -\\
\cdashline{5-16}[.6pt/2pt]

& & \checkmark & & \textbf{58.9} & \textbf{71.8} & \textbf{64.7} & - & - & - & 39.7 & 46.1 & 42.7 & 60.6 & 62.3 & 61.4\\

&  & & \checkmark & 57.1 & 69.8 & 62.8 & - & - & - & \textbf{52.3} & \textbf{61.5} & \textbf{56.5} & \textbf{61.1} & \textbf{86.7} & \textbf{71.7}\\
\hline

\multirow{3}{*}{CE-GZSL}
& \checkmark & & & 69.2 & 88.3 & - & 69.7 & 91.8 & - & 78.7 & 73.5 & - & 68.1 & 89.4 & -\\
\cdashline{5-16}[.6pt/2pt]

& & \checkmark & & 57.5 & 72.3 & 64.1 & 55.5 & 76.7 & 64.4 & 61.1 & 42.8 & 50.3 & 58.3 & 70.8 & 64.0\\

&  & & \checkmark & \textbf{65.7} & \textbf{76.2} & \textbf{70.5} & \textbf{64.6} & 76.7 & \textbf{70.1} & \textbf{66.7} & \textbf{69.1} & \textbf{67.8} & 68.6 & \textbf{80.9} & \textbf{74.2}\\

\hline

\end{tabular}
\caption{The performance comparison of our proposed D$^3$GZSL framework (our), the two-stage classification method based on OOD detection (TS) and idealized version of the two-stage classification method based on OOD detection (IV-TS).}
\label{table:offOD}
\end{table*}

\begin{table}[ht]
\centering
\small

\begin{tabular}{l|p{0.8cm}p{0.7cm}p{0.7cm}ccc}

\hline
Datasets & Baseline & ID$^2$SD & O$^2$DBD & U & S  & H\\

\hline

\multirow{4}{*}{CUB}
& \checkmark & $\times$ & $\times$ & 50.4 & 59.8 & 54.7 \\
& \checkmark & \checkmark & $\times$ & 49.4 & \textbf{63.8} & 55.6 \\
& \checkmark & $\times$ & \checkmark & 51.3 & 59.6 & 55.1 \\
& \checkmark & \checkmark & \checkmark & \textbf{52.3} & 61.5 & \textbf{56.5} \\

\hline

\multirow{4}{*}{FLO}
& \checkmark & $\times$ & $\times$ & 58.6 & 83.3 & 68.8 \\
& \checkmark & \checkmark & $\times$ & \textbf{61.8} & 84.3 & 71.3 \\
& \checkmark & $\times$ & \checkmark & 59.9 & 86.3 & 70.7 \\
& \checkmark & \checkmark & \checkmark & 61.1 & \textbf{86.7} & \textbf{71.7} \\

\hline
\end{tabular}
\caption{The baseline model includes the FG and the classification loss $\mathcal{L}_{cls}$. ID$^2$SD indicates the use of $\mathcal{L}_{be}$ and $\mathcal{L}_{kl}$ losses, while O$^2$DBD represents the use of $\mathcal{L}_{od}$ loss.}
\label{table:ablation}
\end{table}

\begin{table}[ht]
\centering
\begin{tabular}
{c|ccc|ccc}
\hline
\multirow{2}{*}{Method} & \multicolumn{3}{c|}{\textbf{CUB}} & \multicolumn{3}{c}{\textbf{FLO}}\\
& U & S & H &U & S & H\\
\hline
Baseline & 43.7 & 57.7 & 49.7 & 59.0 & 73.8 & 65.6\\

Energy & 49.1 & \textbf{62.3} & 54.9 & 60.3 & \textbf{87.2} & 71.3\\

Softmax & \textbf{52.3} & 61.5 & \textbf{56.5} & \textbf{61.1} & 86.7 & \textbf{71.7}\\

\hline
\end{tabular}
\caption{The performance of our framework is demonstrated under various OOD detection methods, employing f-CLSWGAN as the baseline model in our framework.}
\label{table:oodmethod}
\end{table}

\section{Experiment}
\textbf{Datasets.}
We perform experiments on four ZSL benchmark datasets that are widely used: the Animals with Attributes1\&2 (AWA1 \cite{lampert2013attribute} \& AWA2 \cite{xian2018zero}) dataset, Caltech-UCSD Birds-200-2011 (CUB \cite{wah2011caltech}) dataset, and Oxford Flowers (FLO \cite{nilsback2008automated}) dataset.

\textbf{Evaluation Protocols.}
We evaluate the top-1 accuracy separately on both seen classes ($S$) and unseen classes ($U$) in the generalized zero-shot learning (GZSL). We also use the harmonic mean of these two accuracies ($H = (2 \times S \times U) / (S + U)$) as a performance measure for GZSL.

\textbf{Implementation Details.}
We set the embedding dimension $z$ to 2048 on all datasets.
The classifier $C_s$ outputs logits on all classes, and the classifier $C_o$ outputs logits on seen classes.
The projector $H$ maps softmax probabilities into a two-dimensional space that encodes both ID and OOD information.
The input noise dimension $w$ in the generator is equal to that of the corresponding attributes.
In batch distillation, instances of the same class within a batch serve as positive samples for each other, while those of different class are treated as negative samples.
Here are some of the parameter settings when employing f-CLSWGAN as the baseline model.
We set batch size of 4096 for AWA1, 256 for CUB, 512 for FLO.
The number of generated samples for each unseen category is as follows: 200 for AWA1, 5 for CUB, and 30 for FLO.
We empirically set the loss weights $\lambda = 0.0001$ for AWA1, CUB. We set $\lambda = 0.001$ for FLO.

\subsection{Comparisons with Previous Methods}

In Table \ref{table:sota}, we applied our framework to three previous baseline methods and the DDGAN which is a new  generative method of ZSL we introduced to demonstrate the improvement of our framework on diffusion models.
The results show that we achieved improvements on the AWA1, AWA2, CUB, and FLO datasets.
The most significant improvement was observed with the f-CLSWGAN method, with increases of 3.2\% on the AWA1 dataset, 6.8\% on the CUB dataset, and 6.1\% on the FLO dataset.
The best-performing dataset was CUB, with improvements of 6.8\%, 2.6\%, 2.5\%, and 3.2\% on the four baseline methods, respectively.

Within the CE-GZSL baseline, our $H$ metric delivered the top performance on the AWA1, CUB, and FLO datasets, while ranking second on the AWA2 dataset, only surpassed by DUET \cite{chen2023duet}. Strikingly, when compared to the $S$ metric, our $U$ metric demonstrated a substantial improvement, achieving the highest scores on AWA1, AWA2 and CUB, with 65.7\%, 64.6\% and  66.7\% respectively. It also secured the second-highest performance on the FLO dataset, recording 68.6\%.
The experimental results demonstrate that aligning the distribution of generated samples with that of real samples through out-of-distribution detection is an effective method to address seen bias.

\subsection{Ablation Study}
\subsubsection{Training Strategy Analysis.}
In this section, we compare the outcomes of three different experimental groups. Table \ref{table:offOD} shows the comparison results. The first comprises our proposed one-stage end-to-end training method. The second involves a Two-Stage (TS) classification method based on OOD detection. The third represents an idealized version of the Two-Stage (IV-TS) classification method based on OOD detection. This idealized experiment is designed to simulate the performance of seen and unseen expert classifiers under conditions where the process of OOD detection can classify seen and unseen samples with complete accuracy.
In an ideal scenario, the results of the domain expert classifiers surpass those of our method.
In fact, the OOD detection in TS cannot achieve a 100\% accuracy rate.
Some data that is not within the training distribution is mistakenly assigned to the expert classifiers for classification.
This results in the performance of the TS approach being inferior to that of our proposed method because of the compounding of errors from OOD detection and the expert classifiers (error accumulation).

\subsubsection{Component Analysis.}
Here, we set up an ablation study to examine the impact of various components on our D$^3$GZSL framework.
The baseline model includes the FG and the classification loss $\mathcal{L}_{cls}$.
We conducted three sets of experiments on two benchmark datasets to validate the individual and combined effects of our ID$^2$SD and O$^2$DBD modules.
Through the experimental results presented in Table \ref{table:ablation}, we can draw the following conclusions:
(1) Employing modules ID$^2$SD and O$^2$DBD separately has led to an enhancement in our performance over the baseline method.
(2) When the two modules operate in conjunction, there is a marked enhancement in performance on the $H$ metric.
This suggests that our framework is effective in reducing the discrepancy between the distribution of generated samples and the distribution of real samples. It accomplishes this by optimizing both the in-distribution and out-of-distribution aspects, thereby creating a more cohesive alignment between the two distributions.

\subsubsection{OOD Scoring Strategy Analysis.}
In this paper,  we experimented with two different architectures to verify the impact of using different OOD detection methods on our framework.
We conducted experiments using two methods, Softmax score \cite{hendrycks2016baseline} and Energy \cite{liu2020energy}, on the f-CLSWGAN baseline model.
The experimental results in Table \ref{table:oodmethod} showed that no matter which architecture was used, the performance was significantly improved.
This provides strong evidence for our framework, demonstrating its effective adaptability and scalability, and its ability to be successfully applied to different architectures.

\section{Conclusion}
In this paper, we introduce a generative GZSL framework(D$^3$GZSL) that combines OOD detection and knowledge distillation technologies. Our D$^3$GZSL leverages the OOD detection model to distill the student model, effectively aligning the distribution of generated samples more closely with the distribution of actual samples.
By training a unified classifier as the final GZSL classifier, our framework addresses the issue of accumulated error stemming from two-stage classification in previous ZSL methods based on OOD detection.
Empirical validation through comprehensive experiments demonstrates that our hybrid D$^3$GZSL framework consistently enhances the performance of existing generative GZSL approaches. This novel approach not only leverages cutting-edge techniques but also addresses the limitations of traditional ZSL methodologies, propelling the field towards more precise and reliable zero-shot learning outcomes.

\section{Acknowledgements}
Reported research is partly supported by the National Natural Science Foundation of China under Grant 62176030, and the Natural Science Foundation of Chongqing under Grant cstc2021jcyj-msxmX0568.

\newpage
\bibliography{aaai24}

\end{document}